\documentclass{article}

\usepackage[final,nonatbib]{nips_2018}
\usepackage[square,sort,comma,numbers]{natbib}

\usepackage[utf8]{inputenc} 
\usepackage[T1]{fontenc}    
\usepackage{CJKutf8}        
\usepackage{hyperref}       
\usepackage{url}            
\usepackage{booktabs}       
\usepackage{amsfonts}       
\usepackage{nicefrac}       
\usepackage{microtype}      
\usepackage[super]{nth}

\usepackage{algorithm}
\usepackage{algorithmic}
\usepackage{wrapfig}
\usepackage{enumitem}
\usepackage{subfigure}

\usepackage{amssymb,amsmath}
\usepackage{mathtools}

\title{Deep Learning for Classical Japanese Literature}

\author{
  Tarin Clanuwat\thanks{Corresponding author: \texttt{tarin@nii.ac.jp}, Center for Open Data in the Humanities, Tokyo, Japan.} \\
  Center for Open Data in the Humanities
  \Andhttps://www.overleaf.com/1935325576qsrfhmjzfcvc
  Mikel Bober-Irizar \\
  Royal Grammar School, Guildford
  \AND
  Alex Lamb\\
  MILA, Université de Montréal
  \And
  Kazuaki Yamamoto\\
  National Institute of Japanese Literature
  \AND
  Asanobu Kitamoto \\
  Center for Open Data in the Humanities
  \And
  David Ha \\
  Google Brain
}

\begin{document}
\begin{CJK}{UTF8}{min}

\maketitle

\begin{abstract}

Much of machine learning research focuses on producing models which perform well on benchmark tasks, in turn improving our understanding of the challenges associated with those tasks. From the perspective of ML researchers, the content of the task itself is largely irrelevant, and thus there have increasingly been calls for benchmark tasks to more heavily focus on problems which are of social or cultural relevance. In this work, we introduce Kuzushiji-MNIST, a dataset which focuses on \textit{Kuzushiji} (cursive Japanese), as well as two larger, more challenging datasets, Kuzushiji-49 and Kuzushiji-Kanji. Through these datasets, we wish to engage the machine learning community into the world of classical Japanese literature.

\end{abstract}

\section{Introduction}

Recorded historical documents give us a peek into the past. We are able to glimpse the world before our time; and see its culture, norms, and values to reflect on our own. Japan has very unique historical pathway. Historically, Japan and its culture was relatively isolated from the West, until the Meiji restoration in 1868 where Japanese leaders reformed its education system to modernize its culture. This caused drastic changes in the Japanese language, writing and printing systems. Due to the modernization of Japanese language in this era, cursive Kuzushiji (くずし字) script is no longer taught in the official school curriculum. Even though Kuzushiji had been used for over 1000 years, most Japanese natives today cannot read books written or published over 150 years ago.
\cite{meiji_education,hashimoto2017kuzushiji}

\begin{figure}[!htb]
\vskip -0.05in
\begin{center}
\centerline{\includegraphics[width=0.33\columnwidth]{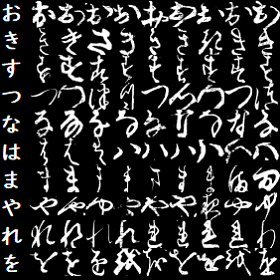}\hspace{1mm}\includegraphics[width=0.66\columnwidth]{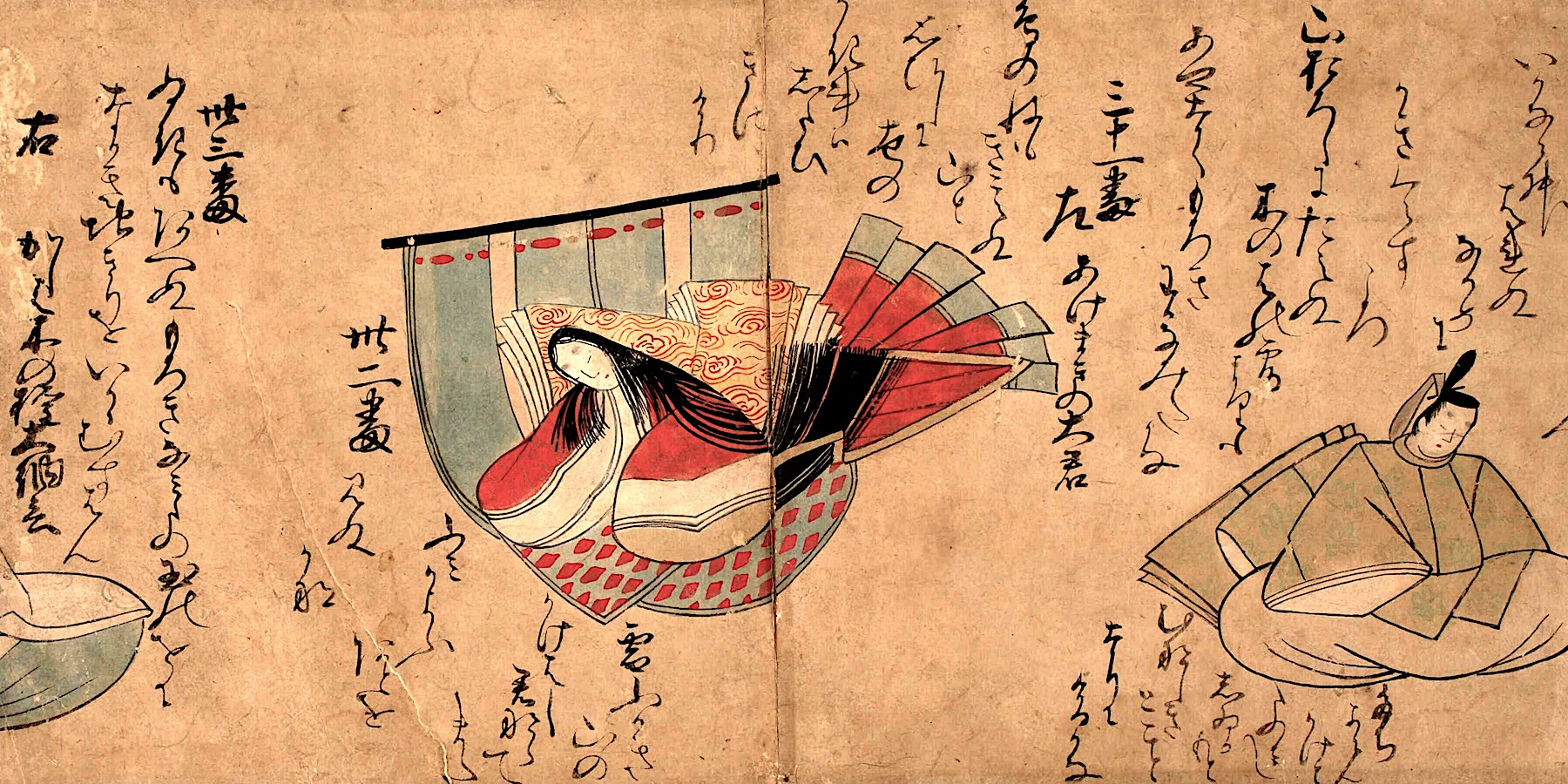}}
\vskip -0.00 in
\caption{Most Japanese cannot read books over 150 years old, written in cursive \textit{Kuzushiji} style. The 10 classes of Kuzushiji-MNIST, first column showing the modern \textit{Hiragana} counterpart (left). Example of a Kuzushiji literature scroll, \textit{Genjimonogatari Uta Awase}『源氏歌合絵巻』\cite{genji_scroll} (right).}
\label{fig:kuzushiji10}
\end{center}
\vskip -0.20 in
\end{figure}

\begin{figure}[!htb]
\vskip -0.15in
\begin{center}
\centerline{\includegraphics[width=0.5\columnwidth]{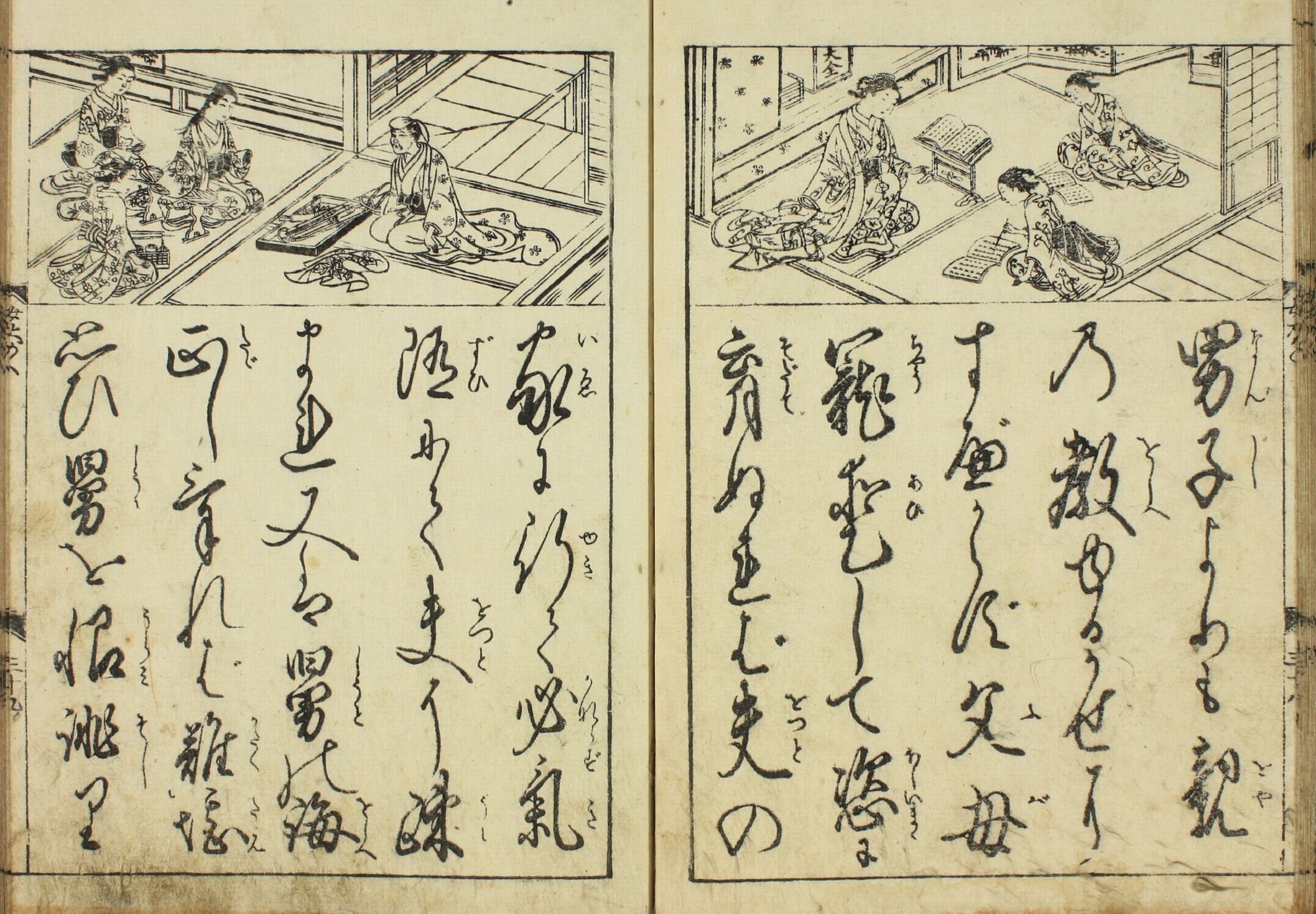}
\hspace{3mm}
\includegraphics[width=0.468\columnwidth]{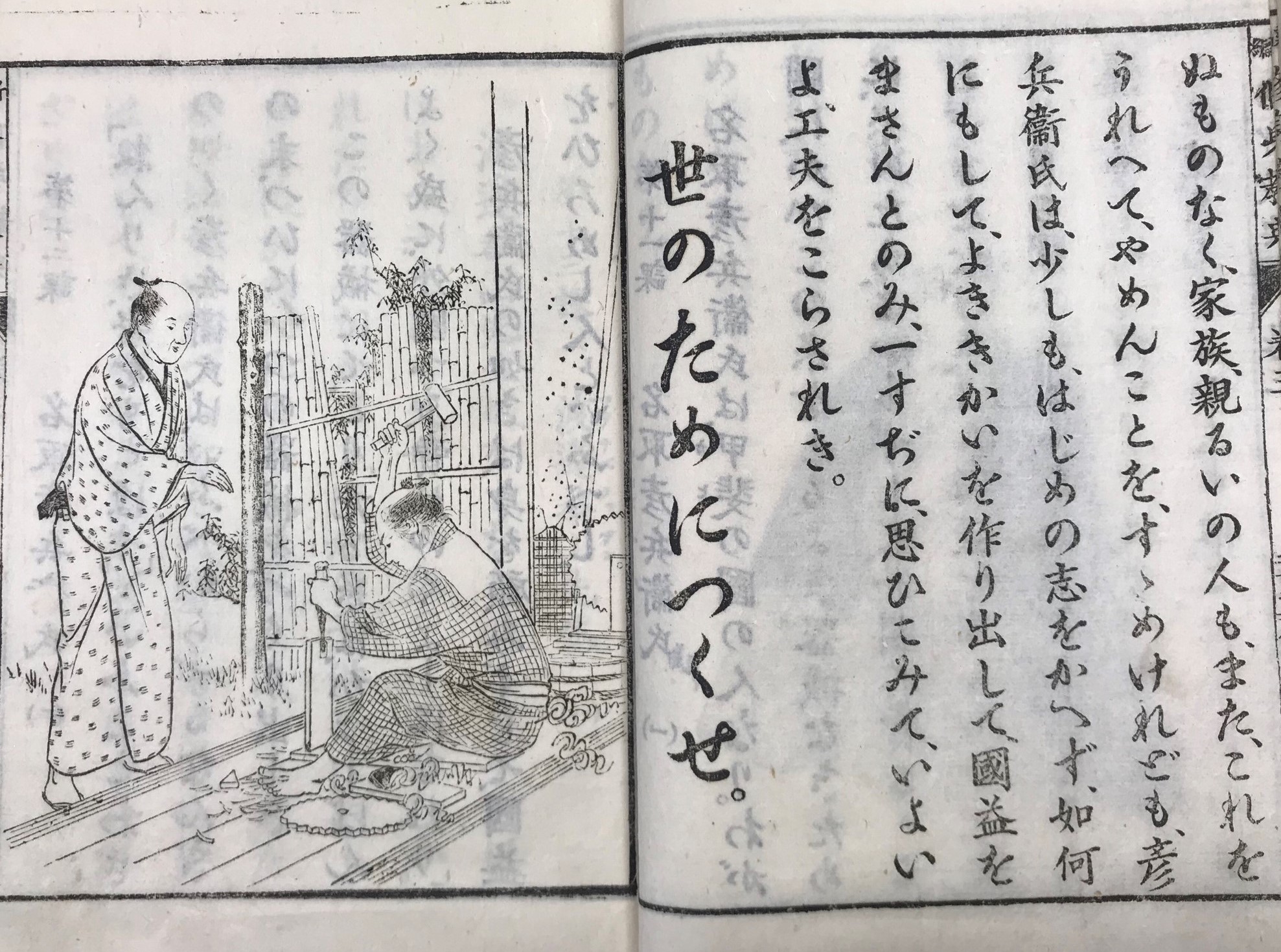}}
\vskip -0.1 in
\caption{The difference between a text printed in 1772 and one printed in 1900. \textit{Onna Daigaku}『女大学』~\cite{textbook_edo} is a book for women in the Edo period (left). \textit{Shinpen Shūshinkyouten Vol.3}『新編修身教典三巻』~\cite{textbook_meiji33} is a textbook right after the standardization of Japanese in 1900 (right).}
\label{fig:textbook}
\end{center}
\vskip -0.35 in
\end{figure}

According to the General Catalog of National Books \cite{national_catalog} there have been over 1.7 million books written or published in Japan prior to 1867. In addition to the number of registered books in the national catalog, we estimate that in total there are over 3 million unregistered books and a billion historical documents preserved nationwide. Despite ongoing efforts to create digital copies of these documents—a safeguard against fires, earthquakes, and tsunamis—most of the knowledge, history, and culture contained within these texts remains inaccessible to the general public. While we have many digitized copies of manuscripts and books, only a small number of people with Kuzushiji education are able to read them and work on them, leading to a huge dataset of Japanese cultural works which cannot be read by non-experts.

\begin{figure}[!htb]
\vskip -0.05in
\begin{center}
\centerline{\includegraphics[width=1.0\columnwidth]{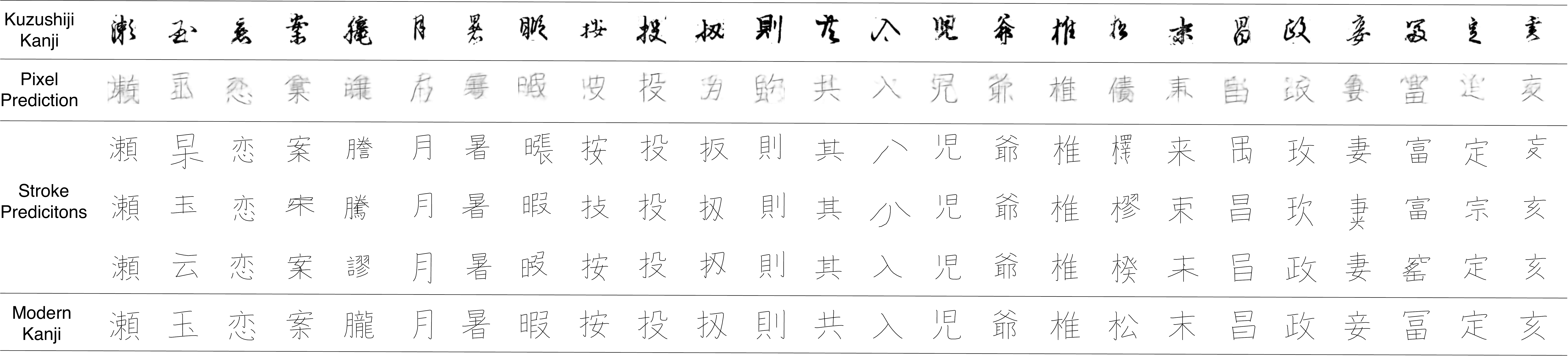}}
\vskip -0.1 in
\caption{Our domain transfer experiment, generating Modern Kanji from the Kuzushiji Kanji for unseen characters. (Section \ref{sec:domaintransfer}).}
\label{fig:kuzushijikanji}
\end{center}
\vskip -0.25 in
\end{figure}

In this paper we introduce a dataset specifically made for machine learning research to engage the community to the field of Japanese literature. In this work, we release three easy-to-use preprocessed datasets: Kuzushiji-MNIST, a dataset which focuses on \textit{Kuzushiji} (cursive Japanese), as well as two larger, more challenging datasets, Kuzushiji-49 and Kuzushiji-Kanji. Kuzushiji-MNIST is designed as a drop-in replacement for the MNIST~\cite{lecun1998mnist} dataset.
In addition, we present baseline classification results on Kuzushiji-MNIST and Kuzushiji-49 using recent models, and also apply generative modelling to a domain transfer task between unseen Kuzushiji Kanji and Modern Kanji (See Figure~\ref{fig:kuzushijikanji}). Through these datasets and experiments, we wish to intoducethe machine learning community into the world of classical Japanese literature.\footnote{Location of dataset with instructions: \url{https://github.com/rois-codh/kmnist}}
\begin{figure}[!htb]
\vskip -0.05in
\begin{center}
\centerline{\includegraphics[width=0.5\columnwidth]{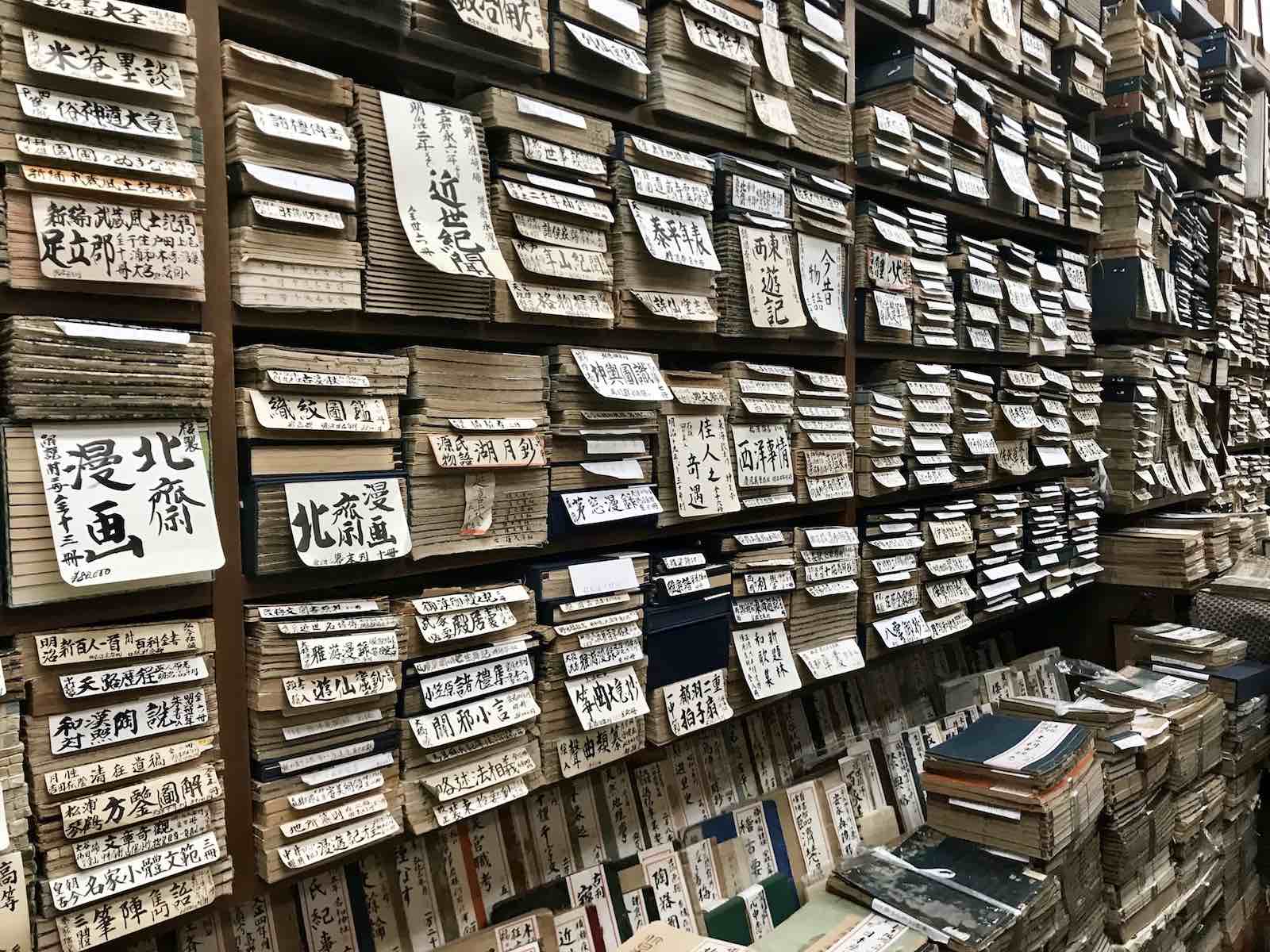}\includegraphics[width=0.5\columnwidth]{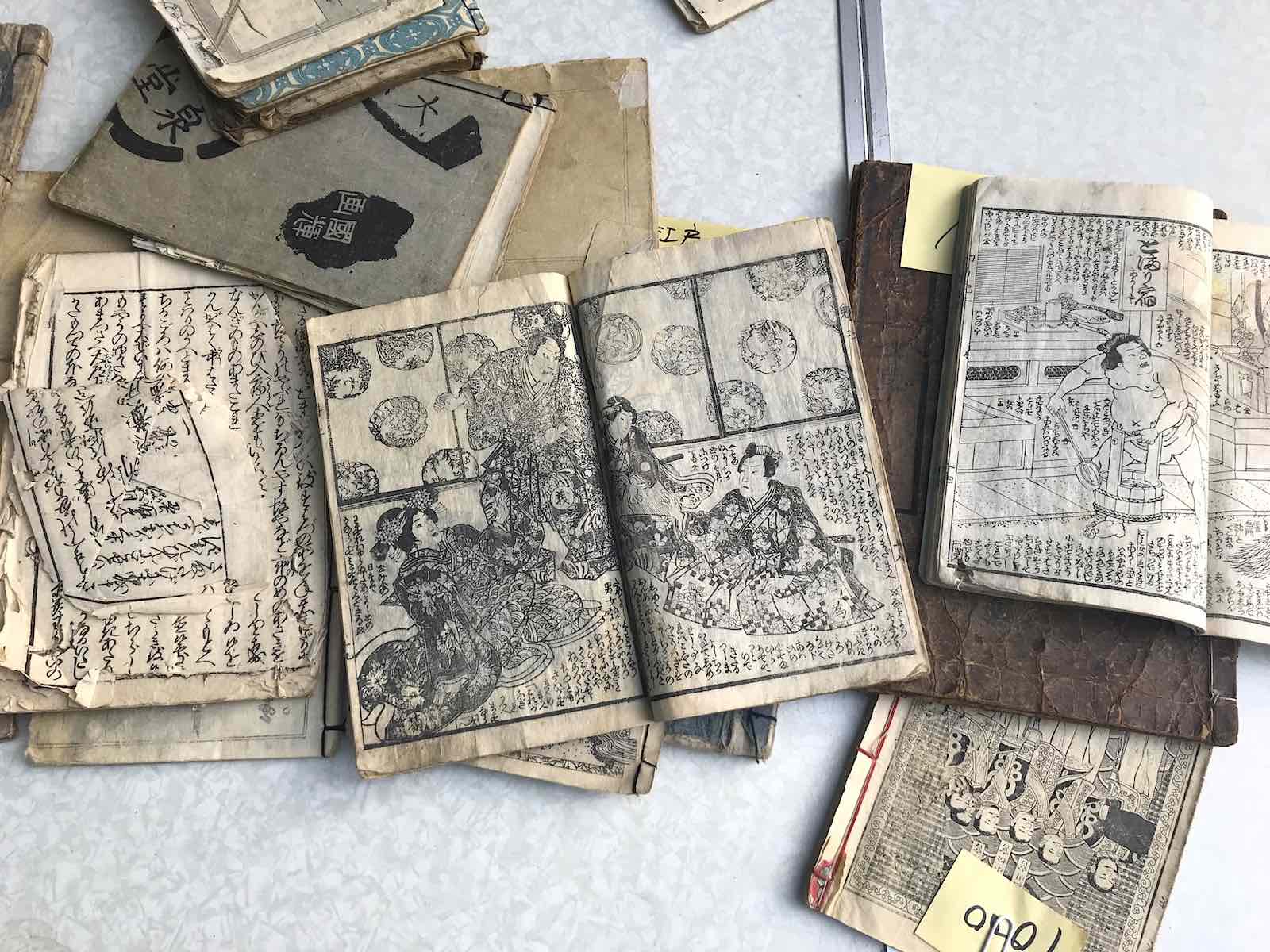}}
\vskip -0.00 in
\caption{In addition to archives, classical books are circulated in manuscript bookstores, online auctions and the annual manuscript auction event held in Jimbocho, Tokyo. Ohya Shobo bookstore in Jimbocho (left). Edo books sold at the Sunday flea market at the Hanazono Shrine, Tokyo (right).}
\label{fig:source_examples}
\end{center}
\vskip -0.30 in
\end{figure}

\section{Kuzushiji Dataset}

The Kuzushiji dataset is created by the National Institute of Japanese Literature (NIJL), and is curated by the Center for Open Data in the Humanities (CODH). In 2014, NIJL and other institutes begun a national project to digitize about 300,000 old Japanese books, transcribing some of them, and sharing them as open data for promoting international collaboration. During the transcription process, a bounding box was created for each character, but literature scholars did not think they were worth sharing. From a machine learning perspective, CODH suggested to make a separate dataset for bounding boxes on a page, because that can be used as the basis for many machine learning challenges and working towards automated transcription. As a result, the \textit{full} Kuzushiji dataset was released in November 2016, and now the dataset contains 3,999 character types and 403,242 characters \cite{codh_pmjt}.

Our hope is that through releasing datasets in familiar formats, we can encourage dialog between the ML and Japanese literature communities. We pre-processed characters scanned from 35 classical books printed in the \nth{18} century and organized the dataset into 3 parts: \textbf{(1)} Kuzushiji-MNIST, a drop-in replacement for the MNIST~\cite{lecun1998mnist} dataset, \textbf{(2)} Kuzushiji-49, a much larger, but imbalanced dataset containing 48 Hiragana characters and one Hiragana iteration mark, and \textbf{(3)} Kuzushiji-Kanji, an imbalanced dataset of 3832 \textit{Kanji} characters, including rare characters with very few samples.

\begin{figure}[!htb]
\vskip -0.05in
\begin{center}
\centerline{\includegraphics[width=0.8\columnwidth]{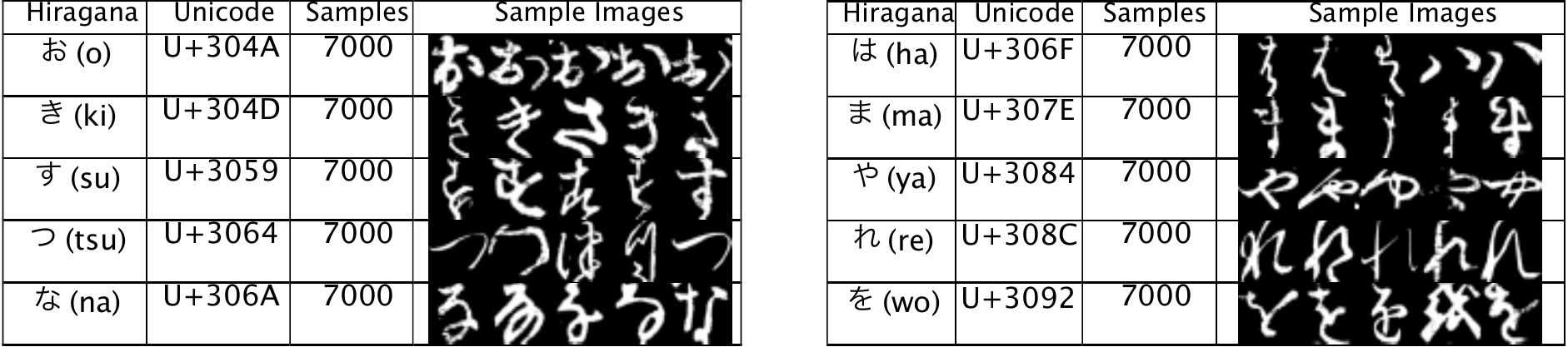}}
\vskip -0.05 in
\caption{The 10 classes of Kuzushiji-MNIST. Train and test set sizes are 6,000 and 1,000 per class.}
\label{fig:kmnist_description}
\end{center}
\vskip -0.25 in
\end{figure}

\begin{figure}[!htb]
\vskip -0.05in
\begin{center}
\centerline{\includegraphics[width=0.15\columnwidth]{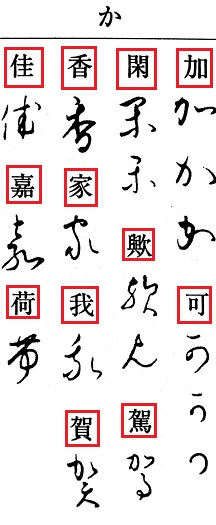}}
\vskip -0.01 in
\caption{The \textit{hentaigana} for Ka (か) can be written using 12 different root characters (\textit{jibo}, in red)~\cite{kuzushijidict}, with some of these root characters themselves having multiple ways of being written. Many of the characters in our datasets have multiple ways of being written, so successful models need to be able to model the multi-modal distribution of each class, making the problem more challenging.}
\label{fig:hentaiganaka}
\end{center}
\vskip -0.25 in
\end{figure}

Since MNIST restricts us to 10 classes, much fewer than the 49 needed to fully represent Kuzushiji Hiragana, we chose one character to represent each of the 10 rows of Hiragana when creating Kuzushiji-MNIST. One characteristic of classical Japanese which is very different from modern one is that Classical Japanese contains \textit{Hentaigana} (変体仮名). \textit{Hentaigana} or \textit{variant kana}, are Hiragana characters that have more than one form of writing, as they were derived from different Kanji. Therefore, one Hiragana class of Kuzushiji-MNIST or Kuzushiji-49 may have many characters mapped to it. For instance, as seen in Figure~\ref{fig:kmnist_description}, there are 3 different ways to write「つ」because this character was derived from different Kanji (川 and 津).

Another example of this many-to-one mapping is shown in Figure ~\ref{fig:hentaiganaka}. Even though Kuzushiji-MNIST was created as drop-in replacement for the MNIST dataset, the characteristics of Hentaigana and Arabic numbers are completely different, and is one reason why we believe the Kuzushiji-MNIST dataset is more challenging than MNIST.

\begin{figure}[!htb]
\vskip -0.05in
\begin{center}
\centerline{\includegraphics[width=1.0\columnwidth]{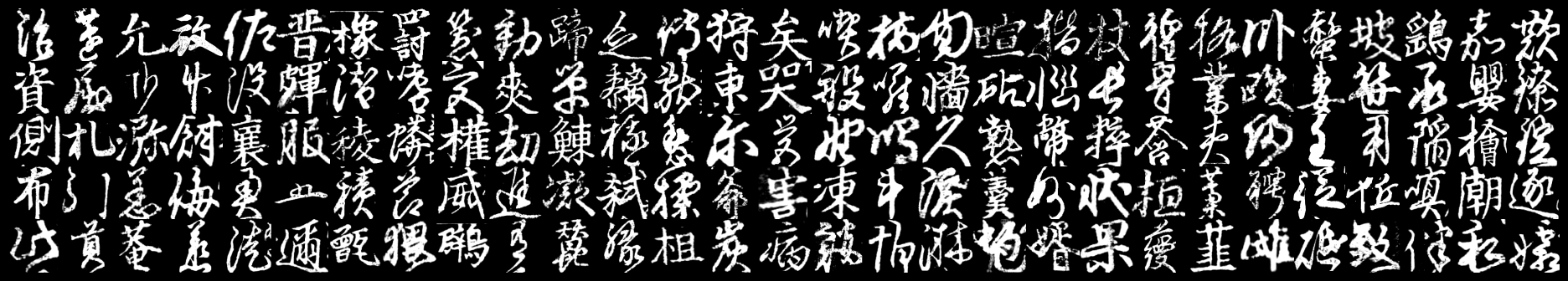}}
\vskip -0.1 in
\caption{Examples of some of the 3832 classes in Kuzushiji-Kanji.}
\label{fig:kanjisample}
\end{center}
\vskip -0.25 in
\end{figure}


The high class imbalance in Kuzushiji-49 and Kuzushiji-Kanji is due to the appearance frequency in the real source books, and kept that way to represent the real data distribution. Kuzushiji-49, as the name suggests, has 49 classes (266,407 images) and Kuzushiji-Kanji has a total of 3832 classes (140,426 images), ranging from 1,766 examples to only a single example per class. Kuzushiji-MNIST and Kuzushiji-49 consist of grayscale images of 28x28 pixel resolution, consistent with the MNIST dataset, while the Kuzushiji-Kanji images are of a larger 64x64 pixel resolution.

\begin{figure}[!htb]
\vskip -0.05in
\begin{center}
\centerline{\includegraphics[width=0.8\columnwidth]{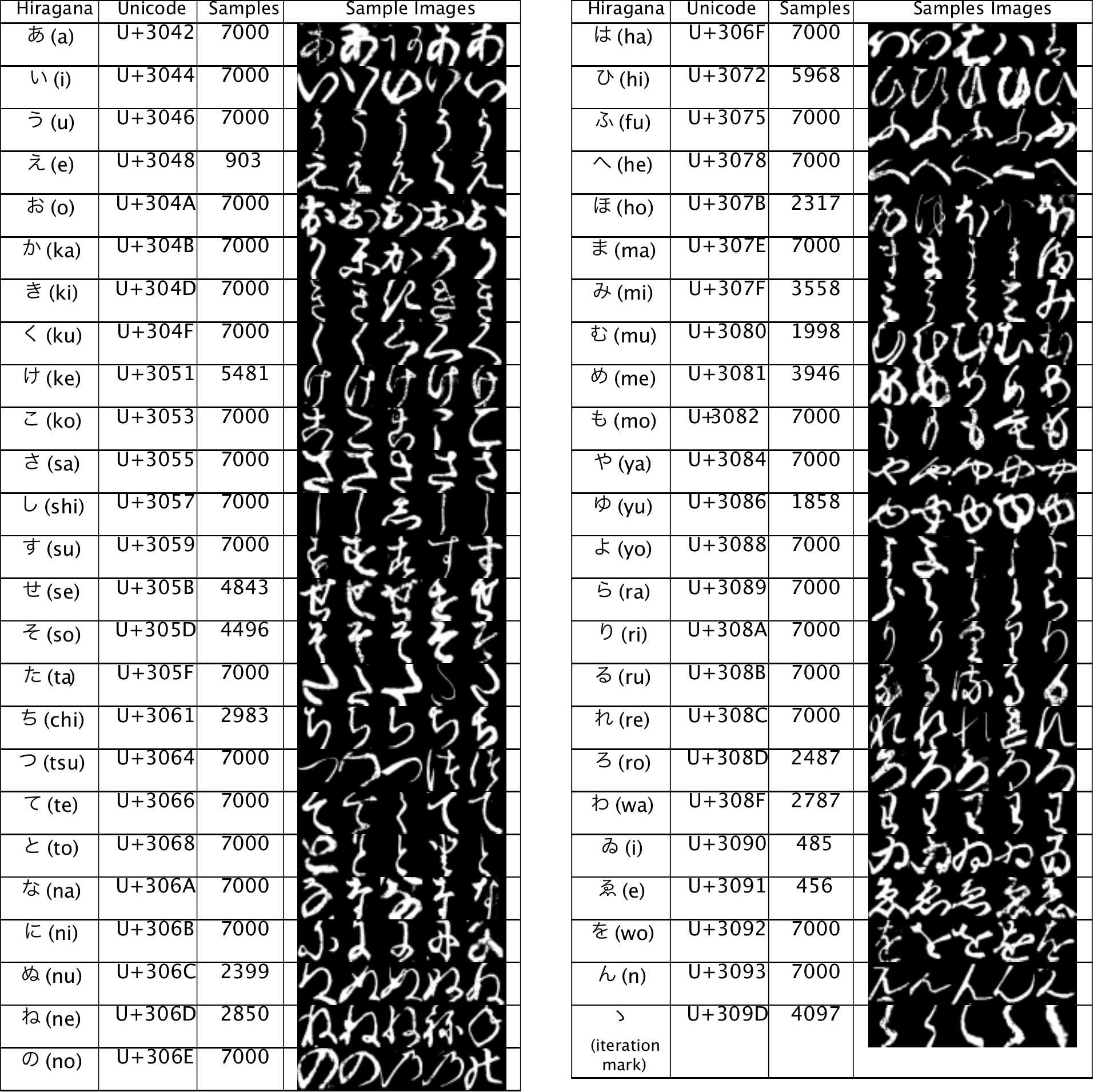}}
\vskip -0.05 in
\caption{Kuzushiji-49 description. Training/Test split is $\frac{6}{7}$ and $\frac{1}{7}$ of each class respectively.}
\label{fig:k_49_description}
\end{center}
\vskip -0.25 in
\end{figure}

In all three datasets, the characters in the train and test sets are sampled from the same 35 books, meaning the data distributions of each class are consistent between the two sets. While Kuzushiji-MNIST is balanced across classes, Kuzushiji-49 has several rare characters with a small number of samples (such as「ゑ」which has only $\sim$ 400 samples).

On the other hand, Kuzushiji-Kanji is a highly imbalanced dataset due to the natural frequency of Kanji appearing in the Kuzushiji literature. In Kuzushiji-Kanji, the number of samples range from over a thousand to only one sample. This dataset is created for more creative experimental tasks rather than merely for classification and character recognition benchmarks.

Our design of a drop-in replacement for MNIST was inspired by the popular Fashion-MNIST~\cite{xiao2017fashion}, a dataset of fashion items that is considerably more difficult than the original MNIST dataset, while maintaining ease of use. One aspect of Fashion-MNIST that we believe decreases model performance compared to MNIST is that many fashion items, such as shirts, T-shirts, or coats look very similar at 28x28 pixel resolution in grayscale, making many samples ambiguous even for humans (Human performance on Fashion-MNIST is only 83.5\%~\cite{xiao2017fashionrepo}). A characteristic of Kuzushiji-MNIST that makes it more difficult compared to MNIST is that there are in fact multiple very different ways to write certain characters, while each way of writing is still unambiguous at 28x28 pixel resolution for human readers, meaning we believe there is less of a performance `cap'. Another difference is that while fashion trends come and go, and what constitute a shirt may be different a hundred years from now, Kuzushiji will always remain Kuzushiji. We believe both Fashion-MNIST and Kuzushiji-MNIST will be useful companions to the original MNIST dataset for the research community.

\section{Experiments}

\subsection{Classification Baselines for Kuzushiji-MNIST and Kuzushiji-49}

\begin{table}[!htb]
\begin{center}
\begin{small}
\begin{tabular}{lccc}
\toprule
Model & MNIST~\cite{lecun1998mnist} & Kuzushiji-MNIST & Kuzushiji-49\\
\midrule
4-Nearest Neighbour Baseline & 97.14\% & 91.56\% & 86.01\% \\
Keras Simple CNN Benchmark~\cite{chollet2015keras} & 99.06\% & 95.12\% & 89.25\% \\
PreActResNet-18~\cite{he2016identity} & \textbf{99.56\%} & 97.82\% & 96.64\% \\
PreActResNet-18 + Input Mixup~\cite{zhang2017mixup} & 99.54\% & 98.41\% & 97.04\% \\
PreActResNet-18 + Manifold Mixup~\cite{verma2018manifold} & 99.54\% & \textbf{98.83\%} & \textbf{97.33\%} \\
\bottomrule
\end{tabular}
\end{small}
\end{center}
\caption{Test set accuracy, computed as mean of per-class accuracies to address class imbalance.}
\label{tab:baseline_accuracies}
\vskip -0.2 in
\end{table}

We present baseline classification results on Kuzushiji-MNIST and Kuzushiji-49 in Table~\ref{tab:baseline_accuracies}. We consider 4 different baselines: A simple 4-nearest neighbours algorithm, a small 2-layer convolutional network, an 18-layer ResNet~\citep{he2016identity}, and a ResNet that incorporates a manifold mixup regularizer~\citep{verma2018manifold}. For the training setup details, please refer to the GitHub repository that contains the dataset. By comparing the performance numbers to the original MNIST dataset using various different approaches, we hope these results will provide a sense of the relative difficulty of our dataset.



\subsection{Domain Transfer from Kuzushiji-Kanji to Modern Kanji}
\label{sec:domaintransfer}

In addition to classification, we are interested in more creative uses of our dataset. While existing work~\cite{isola2017image,liu2017unsupervised,wolf2017unsupervised,bousmalis2017unsupervised} on domain transfer focuses on pixel images, we explore instead the transfer from pixel images to \textit{vector} images, across two different domains. Our proposed model aims to generate Modern Kanji versions of a given Kuzushiji-Kanji input, in both pixel and stroke-based formats.

\begin{figure}[!htb]
\vskip -0.05in
\begin{center}
\centerline{\includegraphics[width=1.0\columnwidth]{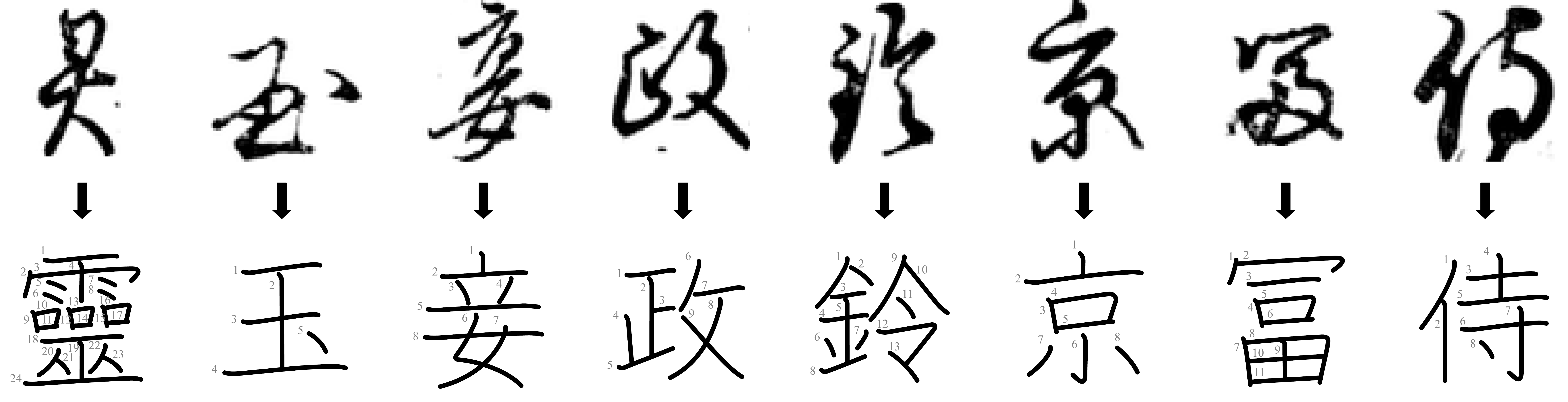}}
\vskip -0.05 in
\caption{Kuzushiji-Kanji 64x64px samples (Top) and stroke-based Modern Kanji versions (Bottom).}
\label{fig:kkanji_problem_statement}
\end{center}
\vskip -0.25 in
\end{figure}

\begin{figure}[!htb]
\vskip -0.05in
\begin{center}
\centerline{\includegraphics[width=1.0\columnwidth]{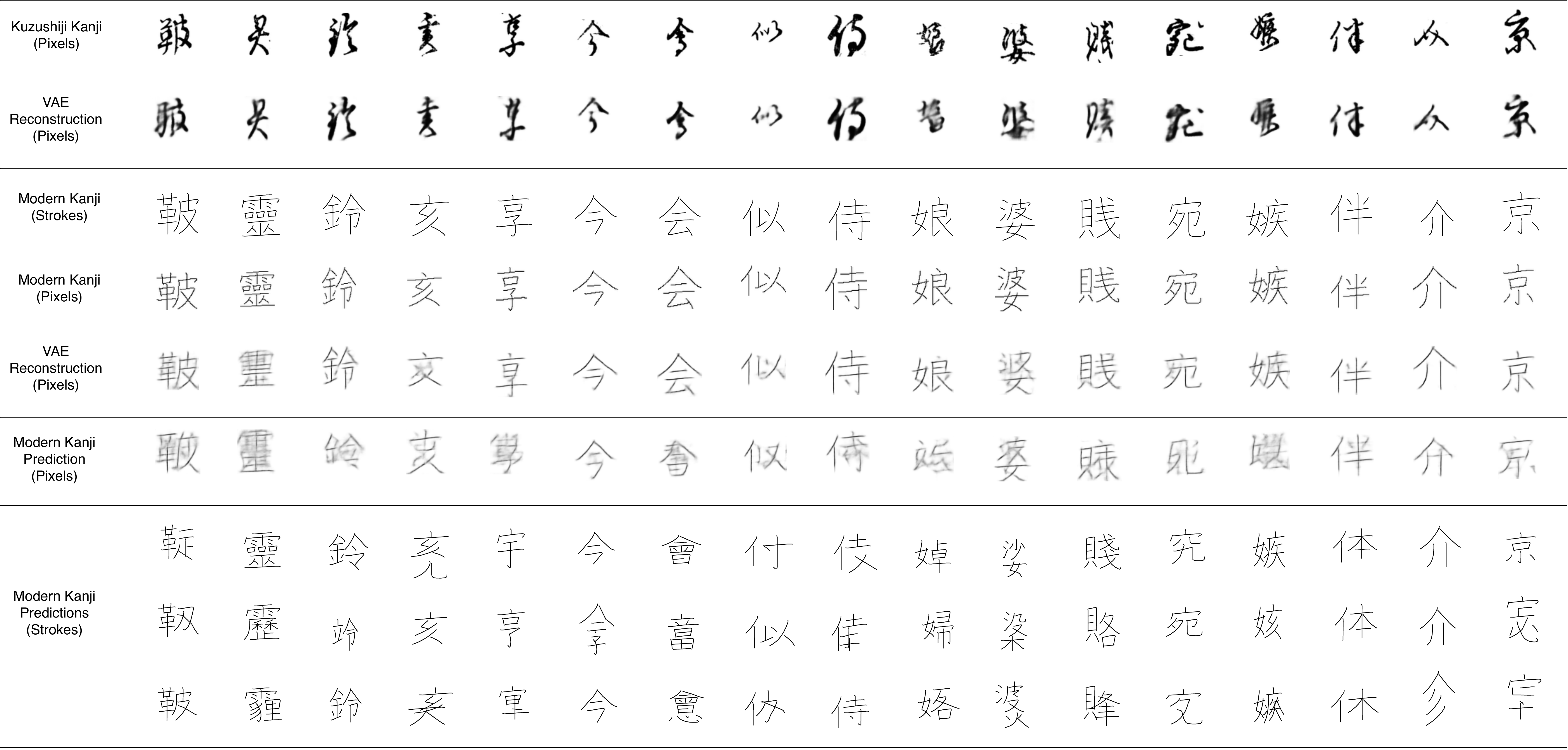}}
\vskip -0.00 in
\caption{More domain transfer examples including the VAE pixel reconstructions for both domains.}
\label{fig:kkanji_results_v2}
\end{center}
\vskip -0.20 in
\end{figure}

We employ KanjiVG~\cite{kanjivg}, a font for Modern Kanji in a stroke-ordered format. Variational Autoencoders~\cite{vae, vae_dm} provide a latent space for both Kuzushiji-Kanji and a pixel version of KanjiVG. A Sketch-RNN~\cite{ha2017neural} model is then trained to generate Modern Kanji strokes, conditioned on the VAE's latent space. Predicting pixel versions of Modern Kanji using a VAE also aids human transcribers as the blurry regions of the output can be interpreted as uncertain regions to focus on. In addition to the earlier Figure~\ref{fig:kuzushijikanji}, see Figure~\ref{fig:kkanji_results_v2} below for a demonstration of our model on test set examples.

\begin{figure}[!htb]
\vskip -0.00in
\begin{center}
\centerline{\includegraphics[width=1.0\columnwidth]{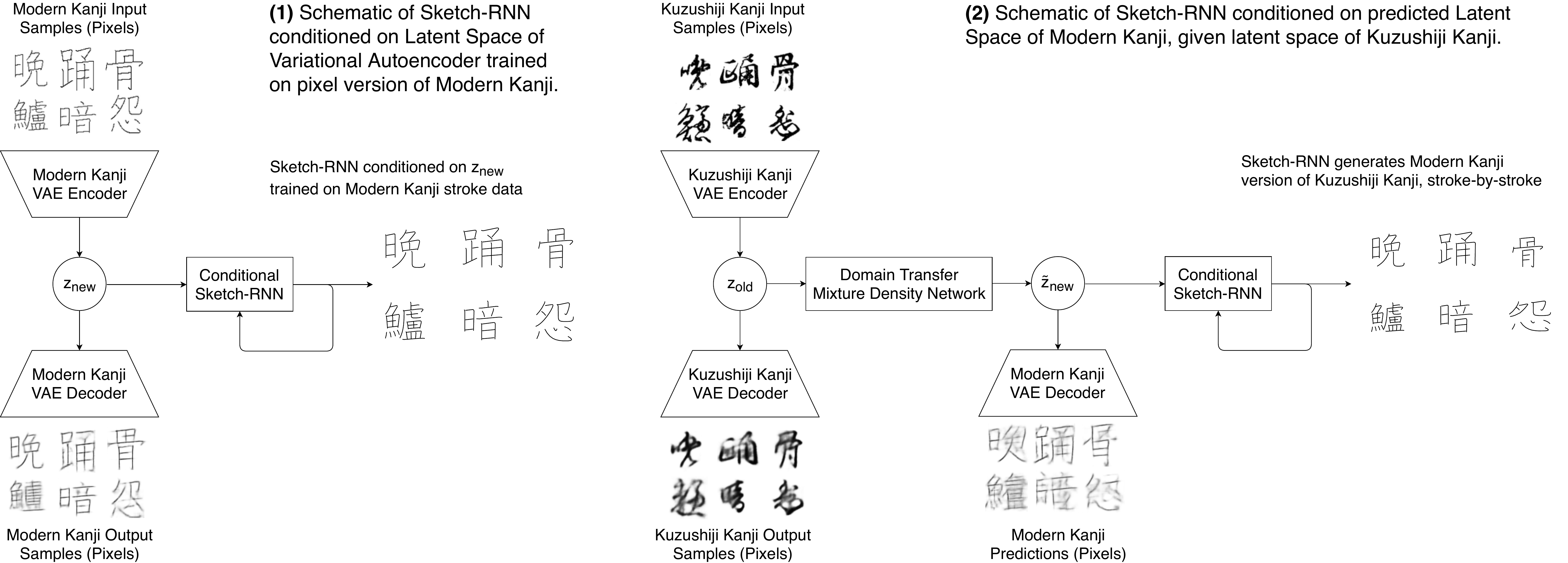}}
\vskip -0.00 in
\caption{Overview of our approach. \textbf{(1)} We first train a VAE on pixel version of KanjiVG (Modern Kanji), and a Sketch-RNN model to generate stroke versions of KanjiVG conditioned on the latent space, $z_{\text{new}}$. \textbf{(2)} We train a VAE on Kuzushiji-Kanji, and train a Mixture Density Network~\cite{bishop_mdn} to predict $P(z_{\text{new}} | z_{\text{old}})$. We generate stroke versions of Modern Kanji based on the predicted $\widetilde{z}_{\text{new}}$.}
\label{fig:kkanji_schematic}
\end{center}
\vskip -0.20 in
\end{figure}

In Figure~\ref{fig:kkanji_schematic}, we present an overall diagram of our approach. We first train two separate Convolutional Variational Autoencoders, one on the Kuzushiji-Kanji dataset, and also a second on a pixel version of KanjiVG dataset rendered to 64x64 pixel resolution for consistency. The architecture for the VAE is identical to \cite{ha2018recurrent} and both datasets are compressed into their own respective 64-dimensional latent space, $z_{\text{old}}$ and $z_{\text{new}}$. As in previous work~\cite{ha2017neural}, we do not optimize the KL loss term below a certain threshold, ensuring some information capacity while enforcing the Gaussian prior on $z$.

\begin{algorithm}[!htb]
\begin{algorithmic}
\begin{small}
\STATE 1. Train two separate Variational Autoencoders~\cite{vae, vae_dm} on pixel version of KanjiVG and Kuzushiji-Kanji.
\STATE 2. Train Mixture Density Network~\cite{bishop_mdn} to model $P(z_{\text{new}} | z_{\text{old}})$ as mixture of Gaussians.
\STATE 3. Train Sketch-RNN~\cite{ha2017neural} to generate KanjiVG strokes conditioned on either $z_{\text{new}}$ or $\widetilde{z}_{\text{new}} \sim P(z_{\text{new}} | z_{\text{old}})$.
\end{small}
\end{algorithmic}
\caption{Summary of training procedure in domain transfer experiment.}
\label{kanji_transfer_training_procedure}
\vskip -0.00in
\end{algorithm}

We then train a Mixture Density Network (MDN)~\cite{bishop_mdn} with 2 hidden layers to model the density function of $P(z_{\text{new}} | z_{\text{old}})$ approximated as a mixture of Gaussians. We can then sample a latent vector $\widetilde{z}_{\text{new}}$ in the domain of Modern Kanji, given a latent vector $z_{\text{old}}$ encoded from Kuzushiji-Kanji. We note that training two separate VAE models on each dataset is much more efficient and achieves better results compared to training a single model end-to-end, which in our experience does not work well, and might explain why previous works~\cite{isola2017image,liu2017unsupervised,wolf2017unsupervised,bousmalis2017unsupervised} require the use of an adversarial loss.

Previous work~\cite{otoro,chinesechar} utilized MDN-RNN to generate stroke-based Chinese characters. In our last step, we train a Sketch-RNN~\cite{ha2017neural} decoder model to generate Modern Kanji conditioned on $\widetilde{z}_{\text{new}}$. There are around 3,600 overlapping Kanji characters between the two datasets. For characters that are not in Kuzushiji-Kanji, we condition the model on the $z_{\text{new}}$ encoded from KanjiVG data to generate the stroke data also from KanjiVG, see (1) in Figure~\ref{fig:kkanji_schematic}. For characters that are in the overlapping 3,600 set, we use the $\widetilde{z}_{\text{new}}$ sampled from the MDN conditioned on $z_{\text{old}}$, to generate the stroke data also from KanjiVG, as per (2) in Figure~\ref{fig:kkanji_schematic}. By doing this, the Sketch-RNN training procedure can fine tune aspects of the VAE's latent space that may not capture well parts of the data distribution of Modern Kanji when trained only on pixels, by training it again on the stroke version of the dataset.

\section{Future Directions}

We believe the Kuzushiji datasets will not only serve as a benchmark to advance classification algorithms, but also contribute to more creative areas such as generative modelling, adversarial examples, few-shot learning, transfer learning and domain adaptation. To foster community building, we plan to organize machine learning competitions using Kuzushiji datasets to encourage further development of these research areas. We are also working on expanding the size of the dataset, and by next year, the size of the full Kuzushiji dataset will expand to over a million character images. We hope these efforts will encourage further collaboration between different research fields and at the same time, help preserve the cultural knowledge and heritage of Japanese history.


\bibliography{main}
\bibliographystyle{abbrv}

\end{CJK}
\end{document}